# Patent Claim Generation by Fine-Tuning OpenAI GPT-2

**Jieh-Sheng Lee** and **Jieh Hsiang**

Department of Computer Science and Information Engineering
National Taiwan University

{d04922013, jhsiang}@ntu.edu.tw

## Abstract

In this work, we focus on fine-tuning an OpenAI GPT-2 pre-trained model for generating patent claims. GPT-2 has demonstrated impressive efficacy of pre-trained language models on various tasks, particularly coherent text generation. Patent claim language itself has rarely been explored in the past and poses a unique challenge. We are motivated to generate coherent patent claims automatically so that augmented inventing might be viable someday. In our implementation, we identified a unique language structure in patent claims and leveraged its implicit human annotations. We investigated the fine-tuning process by probing the first 100 steps and observing the generated text at each step. Based on both conditional and unconditional random sampling, we analyze the overall quality of generated patent claims. Our contributions include: (1) being the first to generate patent claims by machines and being the first to apply GPT-2 to patent claim generation, (2) providing various experiment results for qualitative analysis and future research, (3) proposing a new sampling approach for text generation, and (4) building an e-mail bot for future researchers to explore the fine-tuned GPT-2 model further.

## 1. Introduction

Deep learning and pre-training models have demonstrated excellent results in several language tasks recently. Particularly, fine-tuning the pre-trained models such as ELMo (Embeddings from Language Models) [1], OpenAI GPT (Generative Pre-Training) [2], GPT-2 [3] and BERT (Bidirectional Encoder Representations from Transformers) [4] has become the best practice for state-of-the-art results. GPT-2 is the successor to GPT. Although both GPT-2 and BERT are capable of text generation, Wang and Cho [5] found that GPT-2 generations are of better quality. In fact, GPT-2 is claimed to be so powerful that the risk of its malicious use is high. For this reason, OpenAI decided to keep its largest model (1.5B parameters) closed so that there is more time to discuss its ramifications.

In this work, we generated patent claims by fine-tuning the released 345M medium version [6]. Overall we are impressed by how coherent and complicate the generated patent claims could be, although not all text are generated equally in terms of quality. We are also surprised by how few training steps were necessary for GPT-2 to generate the first text that looks like a patent claim. It is a matter of time that the largest and more powerful model will be released to the public. Therefore, it is better to experiment on GPT-2 and contemplate on its impact on patent research from the beginning of GPT-2 development.

## 2. Related Work

In the patent field, Aristodemou et al. [7] reviewed 57 recent articles on the use of artificial intelligence methods, machine learning and deep learning approaches for analyzing intellectual property data. The analysis is further divided into four main categories: knowledge management, technology management, economic value, and extraction of information. Lupu et al. [8] pointed out that, among patent-related applications, modern neural networks are applied for machine translation primarily and there is a wide open field of opportunities for other tasks such as patent analysis, patent valuation and patent classification. It was also anticipated that the remarkable success of deep learning will certainly be tested on patent data someday.

In the computer science field, NLP (Natural Language Processing) turns text into structured data and NLG (Natural Language Generation) turns structured data back to text. Recently, transfer learning based on Transformer models [9], such as GPT, BERT, and GPT-2, outperformed significantly on various tasks after using pre-trained language models on large-scale corpora. The two-stage framework (pre-training & fine-tuning) is so effective that it is claimed as the arrival of the "ImageNet moment for NLP" [10]. We observed that the success of Deep Learning in NLP field has also spread to NLG field. In this work, we are motivated to apply the latest NLG techniques to the patent field. We see it as an opportunity for patent professionals to generate or extract valuable information from patent data.

## 3. Data

Our training dataset contains 555,890 patent claims of the granted U.S. utility patents in 2013. All of the claims are the first and independent claims. How to train GPT-2 with dependent claims and other independent claims is a topic for future research. We prepared our data from two perspectives: span-based and SQL-based. The former is about splitting a patent claim into shorter text spans. It makes claims easier to comprehend. The latter is about sharing SQL statements for future researchers, instead of sharing conventional raw data. The stages of our data pipeline include (1) raw data collection, (2) claim span identification and tagging, and (3) data encoding for GPT-2. In the rest of the section we first explain our two perspectives, then the data pipeline.

### 3.1. Span-based

Patent claims are longer than ordinary sentences in NLP field. We observed that a lengthy patent claim is usually decomposed into multiple claim spans. A *claim span* is a segment of the claim text. We also observed that, in fact, segmentations exist in human-curated claim text already. The separation of lines in official patent documents is an implicit segmentation. For example, as shown in **Fig. 1**, the first claim of the US9229634B2 patent is decomposed into several claim spans.

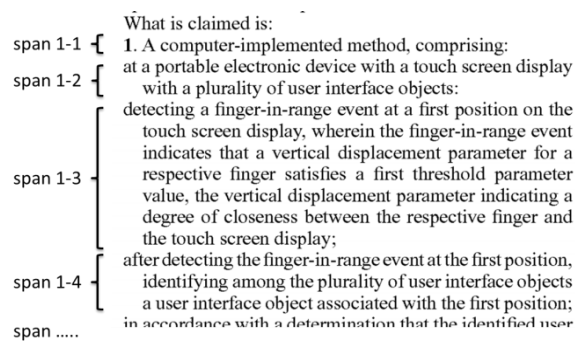

**Fig. 1.** Claim spans of US9229634B2

The purpose of claim spans is two-folded. First, we measure how fast GPT-2 learns from patent corpus by observing how frequent such claim spans are generated. Second, a claim span is a reasonable approximation to an inventive step, component or element for humans to comprehend. In contrast, the granularity of words or phrases in a patent claim is too fine. The granularity of whole claim is too coarse. A claim span is a relatively suitable unit of inventive thought. Such segmentations of claim spans are rich human annotations and were probably never exploited in literatures.

### 3.2. SQL-based

Although raw patent data is available on the USPTO Open Data Portal [11], we found it advantageous to leverage the Google Patents Public Datasets on BigQuery [12] at a higher level. Compared with conventional raw data, two advantages of using SQL statements are: (1) separation of concerns, i.e., the fetching and processing of raw data could be separated, and (2) clarity and flexibility, i.e., SQL statements are precise and easier to customize by different criteria. Usually, if the fetched data in a dataset has been processed and specific to a problem, it is often harder for other researchers to reuse the data for different situations or processing requirements. The SQL statement for our training dataset is listed in the Appendix A.

### 3.3. Data pipeline & special tags

There are three stages in our data pipeline. The first stage is raw data collection based on SQL

statements. The second stage is to split patent claim text into claim spans. The third stage is to encode the claim spans in the required format for GPT-2 to digest. The first stage is simple. The second and the third stages have more implementation details as below.

At the second stage, we implemented the span segmentation on a heuristic basis. As mentioned, the official patent documents contain human-curated line breaks in patent claims. Our heuristic-based implementation is based on the observation that line breaks are often preceded by punctuation. The punctuation mark is either comma or semicolon most of the time. It is conventional to separate a patent claim in this way when filing a patent application. If not, the patent claim is likely to be in such format by human curation before granted.

The punctuation marks alone are not sufficient to split a claim into spans, however. Comma and semicolon appear elsewhere too. Without extra information, it is hard to tell whether a comma acts as a span separator or just an ordinary mark in a claim span. We observed that, in the queried data from the BigQuery, the character return (i.e., line break) between lines was omitted. Combining the omission of character return and the specific punctuation marks makes, it is feasible to implement the span segmentation we need. For example, the character after a comma and semicolon should be a space character, if the punctuation mark is meant to be an ordinary mark instead of a line break. If the character after a comma or semicolon is not a space character, the punctuation mark is presumed to mean a line break. Such a heuristic span segmentation may be not perfect, but it is sufficient to split many patent claims into spans in an approximate sense.

After identifying claim spans, we add a special tag "@@@" as a span separator to each span except for the last span. We follow Woolf's code [13] to add "<|startoftext|>" to the beginning of a patent claim and "<|endoftext|>" to the end. In this way we prepared our training dataset in a specific format. At the inference stage later, we can also identify the beginning and the end of a patent claim in all generated text. After identifying the patent claim, we can further split the generated patent claim into spans based on the span separator.

The third stage of our data pipeline is straightforward. We use the encode function in the GPT-2 code and transform text data into compressed numpy format (*.npz). The format is ready for training iterations and saving time. We shared our training data in both numpy format [14] and plain text format [15]. Future researchers can opt for preparing data from scratch (SQL) or reusing our formatted file for GPT-2.

## 4. Experimental Setup

In this section we explain the computing environment we worked on, the code base we derived from, and the model sizes of GPT-2 to consider.

### 4.1. Pre-trained models

The four GPT-2 model sizes built by OpenAI are 117M, 345M, 762M and 1.5B, in terms of the number of parameters in the neural network. Based on an interim update [16], OpenAI released the larger 345M version as a staged release after the initial 117M model. At the same time, OpenAI shared the 762M and 1.5B versions with selected partners in the AI and security communities who are working to improve societal preparedness for large language models.

In the beginning of our work, we found that the 117M model is sufficient for generating impressive results. Future researchers may start from the small model if computing resource is a constraint. In general, the larger the model is, the better the result becomes. To the limit of the computing environment, our experiments are based on the 345M model in this work.

### 4.2. Colab & GitHub

In terms of computing, we leverage Google Colab [17] for GPU and CPU. Colab is a Jupyter notebook environment that runs entirely in the cloud. Although Colab is free, it has a limit of 12 continuous hours per session. For some of our experiments it is sufficient. For training tasks that require more than 12 hours, we save the TensorFlow checkpoints in training to Google Storage and restore them for continuous training in next session. Manual effort is required to initialize a new session on Colab. Although time consuming, such an almost no cost solution may make it easier for researchers to try different experiments.

The GPU available on Colab is NVIDIA Tesla T4 equipped with roughly 15GB memory available. The memory size is sufficient for fine-tuning all layers of the small model (117M), but it is not sufficient for the medium model (345M). A public workaround is to use a memory efficient gradient technique so that it works on Colab. We followed this approach based on Shepperd's repository [18]. The repository is also the same code base for us to fine-tune the pre-trained model with patent claims.

It is noted that TPU (Tensor Processing Unit), more powerful than GPU, is also available on Colab. However, during our experiments, the public TensorFlow-based repositories work with GPU only. A popular PyTorch-based implementation of GPT-2 [19] works with GPU only because the latest official release of PyTorch does not support TPU. It was anticipated that both TensorFlow-based and PyTorch-based repositories will work on TPU soon. If TPU is available, one of our follow-up tasks will be building a pre-trained model from scratch and from patent corpus only. A patent-specific pre-trained model could be an important building block for downstream patent tasks. For example, it would be interesting to know whether fine-tuning such a patent-specific model with CPC information can make patent classification task perform better or generate better patent claims. It would also interesting to know whether fine-tuning the pre-trained with scientific papers can generate or discover relevant patent claims.

## 5. Experiments

We set four goals in our experiments. The first goal is to understand how fast the GPT-2 model adapts to patent claims. We approached this goal by observing the generated text in each step of early fine-tuning. Our second goal is to record the loss values during fine-tuning and observe their converging trend. Our third goal is to analyze the overall quality of the generated patent claims by unconditional random sampling. We devised a new sampling approach in pursuit of higher text quality. Our final goal is to compare the generated patent claims based on different text inputs for conditional random sampling. We further built a prototype of e-mail bot for future researchers to query the GPT-2 model and review its text generation.

### 5.1. How fast GPT-2 adapts

We measured how fast the GPT-2 adapts to patent claims by counting the occurrences of our special tags ("<|startoftext|>", "<|endoftext|>" and "@@@"") in generated text. It is a reasonable expectation that GPT-2 can generate more patent-like text if the number of fine-tuning step increases. It is to our surprise how few steps are required to generate the first patent-like text. As early as at the 36th step, GPT-2 generated the following text and the positions of special tags are also appropriate:

*<|startoftext|> A hand-held mobile device comprising: @@@ a second hand with an articulated index finger comprising a first-stage rotator, a second-stage rotator, said second-stage rotator being rotatable upward and downward along a first pivot point, the apparatus extending through said first pivot point, said second-stage rotator being rotatable downward and upward towards said first pivot point, @@@ said first rotator being rotatable on a first end of said first pivot point when the apparatus is in said first-stage rotator position to provide an at least a second-stage rotator that rotates around a first pivot point; @@@ further, the apparatus has a first pivoting axis that rotates the pivot point; @@@ and the second hand pivoting axis has a second-stage rotator that rotates around a second pivot point, the apparatus rotating the at least a first-stage rotator of said first pivot point when said second-stage rotator is in said first-stage rotator position to provide said at least a second-stage rotator that rotates around a second pivot point. <|endoftext|>*

It is unknown to us how many or how few patents were included in the original 40G WebText that the GPT-2 pre-trained models were trained on. If there were many, it would be still surprising to see the effectiveness of fine-tuning at a very early stage. If there were only a few, the effectiveness would be unreasonable. We leave this doubt to be clarified in the future.

For archiving details, we collected the generated text in the first 100 steps of fine-tuning. The collection is made available online as research data for future study [20]. Statistically the occurrences of the three special tags in the first 100 training steps are shown in **Fig. 2**. It is noted that not all of the occurrences make good sense. The chart is nevertheless a simple and intuitive way to suspect what might be happening in the black box.

A neural network is often a black box in the sense that even a simple neural network with a single hidden layer could be hard to understand. Interpreting black box models has been a challenge in Deep Learning for a long time. Compared with other neural network models, there is a chance that the attention mechanism in Transformer models may provide better interpretability. For example, Vig [21] presented an open-source tool for visualizing multi-head self-attention in Transformer-based language models. The tool is capable of visualizing attention at three levels of granularity (attention-head level, model level and neuron level). It might provide more insights for understanding the first 100 steps from inside out. We leave this a research topic to the future.

In our experiment, we built a baseline and kept the hyperparameters that are common in public repositories, specifically learning rate as *1e-4*, temperature as *1.0*, *top_k* as *40* and batch size as *1*. We ever tried *1e-5* as the learning rate, but the convergence is too slow. Nothing like a patent claim was generated in the first 100 steps, if the learning rate is *1e-5*.

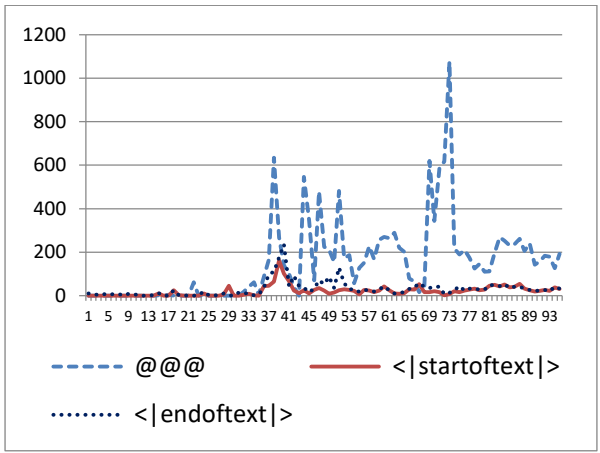

**Fig. 2.** Number of generated special tags in the first 100 steps of fine-tuning

### 5.2. Training loss in fine-tuning

In this experiment we used the same common hyperparameters but changed the learning rate from *1e-4* to *1e-5*. We expected a lower training loss. The convergence of the training losses is shown in **Fig. 3**. A lower learning rate leads to a slower convergence in general. In our case, it took about 18 days on Colab to run 521K training steps. Based on the trajectory, it is reasonable to us that the training loss is likely to decrease after more training steps. At which step it will become flat is unknown, though. How to use a different dataset to validate and prevent overfitting is also unknown. We leave this kind of topics to the future after having more computing resources.

### 5.3. Unconditional random sampling

In this experiment, we explored different approaches of unconditional sampling for patent claim generation. The original GPT-2 used the *top_k* random sampling with a default value *40*. It means sorting by probability and zero-ing out anything below the 40th token when sampling. Holtzman et al. [22] pointed out that a potential issue is the fixed threshold *k* which may be not the best all the time. The quality of generated text depends on the distribution of reasonable words in actual cases. If there are many words one could sample from reasonably, the number *k* might be too low. If there are only a few reasonable words to sample from, the number *k* might be too high. To solve this problem and have a dynamic number of samples, the authors proposed a *top_p* sampling called nucleus sampling. The essence of the *top_p* sampling is that the number of samples depends on zero-ing out anything below the cumulated distribution *p*.

In this work, we propose a different cut-off approach called *dynamic_kp*. The cut-off threshold is based on the relative scale of probability compared with the probability of the top token. For example, in our experiment, we set the cut-off probability as *0.1* of the probability of the top token. We assumed that one order of magnitude is a significant boundary for cut-off and zero-ing out anything below. It has an effect to make the *k* value dynamic in *top_k* sampling and the *p* value dynamic in *top_p* sampling. We provide our source code of the *dynamic_kp* sampling as below. Whether the threshold *0.1* for *dynamic_kp* is the best is up to future experiments. A possible implementation is to provide an interactive interface for users interrogate GPT-2. The best threshold for *dynamic_kp* might be learned from user interactions.

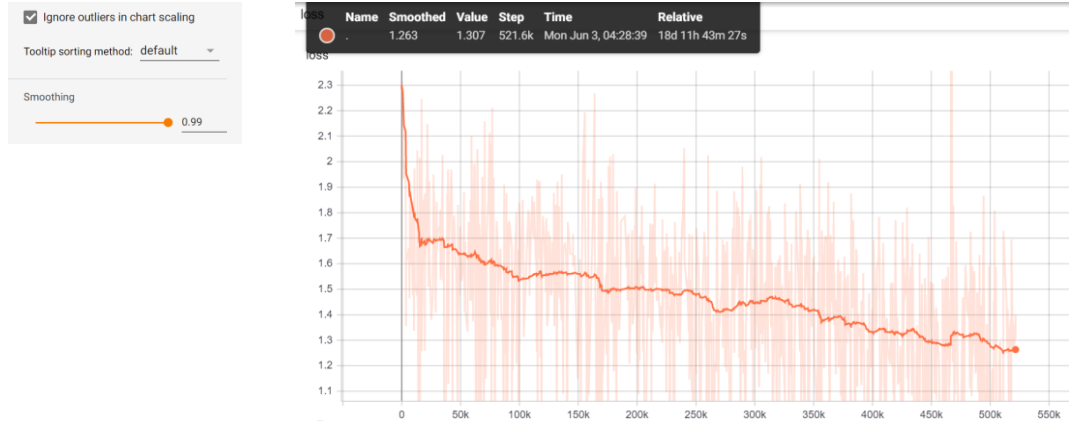

**Fig. 3.** Training loss in fine-tuning

```
def dynamic_kp(logits):
  k=100 # sufficient to identify cut-off

  probs_logits=tf.nn.softmax(logits)
  k_probs,_=tf.nn.top_k(probs_logits,
k=k)
  k_probs=tf.squeeze(k_probs)
  # top 100 probabilities

  probs_max=tf.reduce_max(k_probs)
  # max probability

  k_threshold=tf.multiply(probs_max,0.1)
  probs_mask=tf.to_int32(k_probs>=k_thre
shold)
  num_of_k=tf.count_nonzero(probs_mask,
dtype=tf.int32)
  # the number of tokens above cut-off

  # leverage original top_k code
  values,_=tf.nn.top_k(logits,
k=num_of_k)
  min_values=values[:,-1, tf.newaxis]
  return tf.where(logits < min_values,
tf.ones_like(logits, dtype=logits.dtype)
*-1e10,logits)
```

For comparing different sampling results, we tried *dynamic_kp (0.1)*, *top_k (40)* and *top_p (0.9)* to generated 30 patent claims for each type. Without any cherry-picking, a complete list of the 90 patent claims is archived online as research data for review [23]. We conducted our qualitative analysis based on these 90 samples. Take the following as a positive example:

*A method for a mobile station to communicate with a base station in a wireless communication system, the method comprising:*

*transmitting, to the base station, a first request to enter a high power state, wherein the first request is received according to an operating state;*

*receiving, from the base station, a second request to enter the high power state, wherein the second request is received according to a standby state;*

*determining whether the mobile station is in the standby state; and*

*entering the high power state upon determining that the mobile station is in the standby state.*

We observed that the above generated claim seems coherent, making some practical sense, and having no obvious syntactical or semantical error. Overall we observed a significant number of generated claims having similar and acceptable quality. Among those claims, we also observed a plausible range of diversity and correct sequences of bullet items.

In contrast, the number of samples with poorer quality is also significant. Particularly, some claims are too long and hard to understand. The details of a claim may diverge and end up very far. Sometimes a term or phrase may be even repetitive or obviously incorrect. How to fix these quality problems is a future research topic. For example, regarding the lengthy claims, it might be possible to split them into independent and dependent claims by a downstream task. Or, such an issue might be mitigated (or become worse) after having dependent claims included in the training dataset. For brevity, we selected a few claim spans of poorer quality as below. Interested readers could investigate the complete list online for details.

*……*
*wherein one or more of the control signal sets are used to generate a plurality of image display,*
*wherein one or more of the control signal sets are used to generate a plurality of new image display,*
*……*
*receiving, at the first wireless access point, a third data packet from the second wireless access point, the third data packet sent from the second wireless access point.*
*……*
*e) generating the sequence of data packets from the sequence of data packets based on the selected time for each data packet; and*
*f) transmitting the sequence of data packets, the sequence of data packets having a time and the stored time.*
*……*

*5.4. Conditional random sampling*

In this experiment we followed the same settings and tried conditional random sampling with two different inputs: (1) A deep learning method for patent analysis, (2) A deep learning method for drones. We generated 30 patent claims for each of the *dynamic_kp*, *top_k* and *top_p* samplings. Without cherry-picking, the 90 generated patent claims for both inputs are archived online as research data respectively [24] [25]. Most of our observation is similar to our qualitative analysis on unconditional random sampling and omitted here. It is noted that the text generation quality depends on the input text, however. For example, if the input is longer, it is generally harder for all of the generated text to stay relevant to the input text. If the input is shorter and looks like the beginning of a claim, the text generation quality is usually better. We leave empirical study on quantitative analysis to the future and provide two positive examples to show some reasonable quality of text generation:

*(1) A deep learning method for patent analysis , comprising the steps of:*
*generating a plurality of patent scores for each patent of a plurality of patents in a training set of patent scores, each patent of the plurality of patents having a patent score for each of the plurality of patents in the training set;*
*receiving a patent score for each of a plurality of patents in a training set of patents scores;*
*generating a plurality of patent score differences for each of the plurality of patents, wherein a first patent score difference is generated for a first patent of the plurality of patents based on the received patent score of the first patent and a second patent score difference is generated for a second patent of the plurality of patents based on the received patent score of the second patent;*
*comparing the first patent score difference and the second patent score difference; and*
*generating a patent score for each of the plurality of patents based on the comparison.*

*(2) A deep learning method for drones , comprising the following steps:*
*a. creating an initial base grid and a final base grid by calculating a first total number of points and a first distance between the final base grid and the initial base grid;*
*b. setting up a first grid with a plurality of cells;*
*c. setting up a second grid with a plurality of cells;*
*d. setting up a third grid with a plurality of cells, wherein each cell of the second grid is connected to each cell of the third grid;*
*e. calculating a plurality of total distance durations for each cell in the second grid and the third grid;*
*f. calculating a plurality of total distance durations for each cell in the first grid and the second grid; and*
*g. calculating a plurality of total distance durations for each cell in the final grid and the first grid.*

These two patent claims are very different, even though the majority of the input characters are the same. We observed that the possible details of "patent analysis" and "drones" were generated respectively with acceptable quality. It could be noted that the length of the input is short. The length of output is comparatively long. Therefore, we contemplate on a use case in which, if an inventor is just exploring new ideas and has no whole picture in mind yet, claim generation like this may be a way of augmented inventing. If the speed of GPT-2 inference is fast enough in the future, it should be possible to build an interactive patent drafting assistant. The assistant can suggest next terms, phrases, claim spans or even new ideas. This may open a new window for both qualitative and quantitative analysis on patent claim generation, too. For example, by measuring the gap between user's actual next word and the probability distribution of candidate words in GPT-2, it is possible to measure the accuracy of inferencing. It is also possible to compare the accuracy of different sampling approaches. Such user interaction and quantitative metrics may shed more lights on understanding Transformer models deeper.

## 5.5. ai.patent.bot@gmail.com

The above email address is accessible for testing the fined-tuned GPT-2 model, as long as we can maintain the required computing resource. For testing unconditional random sampling, one can send an email with empty mail body. The mail subject does not matter. For testing conditional random sampling, the mail body should contain the seed text for GPT-2 to inference upon. If the seed text is too long, the encoded tokens within the first half of context window of the model will be reserved for inference and the rest will be discarded. The purpose is for having sufficient space for text generation.

Out of curiosity we sent an email with "Patent claim generation" as mail body to the e-mail bot. A significant percentage of the generated text does not make any sense. Nevertheless, it did generate something interesting like the following. It is noted that the generation quality by our *dynamic_kp* sampling is slightly better than *top_k* and *top_p* in this test. Which one is always better is not conclusive yet.

*Patent claim generation apparatus that generates a patent claim in a first language, comprising:*

*a first-language-specific attribute extraction portion that extracts a first-language-specific attribute from a first patent document;*

*a second-language-specific attribute extraction portion that extracts a second-language-specific attribute from a second patent document;*

*a determination portion that determines, based on a first attribute of the first-language-specific attribute extracted by the first-language-specific attribute extraction portion and a second attribute of the second-language-specific attribute extracted by the second-language-specific attribute extraction portion, whether the first-language-specific attribute and the second-language-specific attribute are identical to each other;*

*an attribute-identification portion that, when the determination portion determines that the first-language-specific attribute and the second-language-specific attribute are identical to each other, identifies a third attribute of the first-language-specific attribute and a fourth attribute of the second-language-specific attribute as matching attributes of the patent document, based on attribute information of the first-language-specific attribute and the second-language-specific attribute; and*

*a patent document determination portion that determines whether the patent document is a valid patent document, based on the third attribute and the fourth attribute identified by the attribute-identification portion..*

## 6. Looking forward

Transformer-based models are at the early stage of development in the Deep Learning field. It won't be surprising if the next version of GPT-2 or BERT sets a new state of the art or the model size increases further. In this section, we look forward briefly some recent efforts on Transformer-based models and their implications on patent claim generation in the future.

First, patent classification is a kind of supervised knowledge which can be leveraged. Training a pre-trained Transformer model with supervised knowledge may outperform the original model. For example, Li, et al. [26] investigated a transferable BERT training framework, which can transfer not only general language knowledge from large-scale unlabeled data but also specific kinds of knowledge from various related supervised tasks, such as next action prediction and sentiment classification. We expect that such a three-stage approach can generate better patent claims if the patent classification information can be learned into the model.

Second, patent claim generation in languages other than English is another line of work. One possibility is to fine-tune a pre-trained model in a different language. Another possibility is to fine-tune a pre-trained multilingual model. The latter is compelling because Pires et al. [27] showed that a multilingual BERT model is able to perform cross-lingual generalization well. This means that the annotations in one language can be used to fine-tune the model for another language. We conjecture that the supervised knowledge, such as patent classification in other languages, can make multilingual patent claim generation more effective.

## 7. Conclusion

We demonstrated that GPT-2 might be a viable solution for augmented inventing. By fine-tuning its 345M model, we contributed hundreds of generated patent claims as research data and provided an e-mail bot for future researchers to experiment with. In our experiences, the emergence of Transformer models such as GPT-2 is a paradigm shift and a tremendous opportunity for patent researchers. We successfully identified a unique language structure in patent claims and applied the GPT-2 model to generate patent claims of reasonable quality. Our qualitative analysis shows promising potentials for future research, such as fine-tuning a larger pre-trained model, or building a pre-trained model from scratch and from patent corpus only. Leveraging more human annotations, such as patent classification, is also a potential way to push the current quality of patent claim generation further. In summary, our work might be a step toward an era of human-machine co-inventing.

## References

[1] M.E. Peters, M. Neumann, M. Iyyer, M. Gardner, C. Clark, K. Lee, L. Zettlemoyer, Deep contextualized word representations, (2018). https://arxiv.org/abs/1802.05365 (accessed April 10, 2018).

[2] A. Radford, K. Narasimhan, T. Salimans, I. Sutskever, Improving Language Understanding by Generative Pre-Training (transformer in real world), (n.d.) 1–12.

[3] A. Radrof, J. Wu, R. Child, D. Luan, D. Amodei, I. Sutskever, Language Models are Unsupervised Multitask Learners, (2018).

[4] J. Devlin, M.-W. Chang, K. Lee, K. Toutanova, BERT: Pre-training of Deep Bidirectional Transformers for Language Understanding, in: Proc. 2019 Conf. North Am. Chapter Assoc. Comput. Linguist. Hum. Lang. Technol. NAACL-HLT 2019, Minneapolis, MN, USA, June 2-7, 2019, Vol. 1 (Long Short Pap., 2019: pp. 4171–4186. https://aclweb.org/anthology/papers/N/N19/N19-1423/.

[5] A. Wang, K. Cho, BERT has a Mouth, and It Must Speak: BERT as a Markov Random Field Language Model, (2019). http://arxiv.org/abs/1902.04094 (accessed March 1, 2019).

[6] OpenAI, GPT-2 source code, (n.d.). https://github.com/openai/gpt-2 (accessed June 2, 2019).

[7] L. Aristodemou, F. Tietze, The state-of-the-art on Intellectual Property Analytics (IPA): A literature review on artificial intelligence, machine learning and deep learning methods for analysing intellectual property (IP) data, World Pat. Inf. 55 (2018) 37–51. doi:10.1016/j.wpi.2018.07.002.

[8] M. Lupu, Information retrieval, machine learning, and Natural Language Processing for intellectual property information, World Pat. Inf. 49 (2017) A1–A3. doi:10.1016/j.wpi.2017.06.002.

[9] A. Vaswani, N. Shazeer, N. Parmar, J. Uszkoreit, L. Jones, A.N. Gomez, L. Kaiser, I. Polosukhin, Attention Is All You Need, (2017). http://arxiv.org/abs/1706.03762 (accessed December 24, 2018).

[10] S. Ruder, NLP's ImageNet moment has arrived, (n.d.). http://ruder.io/nlp-imagenet/.

[11] USPTO, USPTO Open Data Portal, (n.d.). https://developer.uspto.gov/.

[12] Google, Google Patents Public Datasets on BigQuery, (n.d.). https://console.cloud.google.com/bigquery?p=patents-public-data.

[13] M. Woolf, gpt-2-simple, (n.d.). https://github.com/minimaxir/gpt-2-simple.

[14] gpt2-claims-2013_for_345M.npz, (n.d.). https://data.mendeley.com/datasets/b8853hnj7b/draft?a=b99308ff-c24b-428c-96d7-d851962a2714.

[15] gpt2-claims-2013.txt, (n.d.). https://data.mendeley.com/datasets/9dvny7cgcz/draft?a=6ba92bff-b464-4665-90c9-8e03f1ba4a13.

[16] OpenAI, Better Language Models and Their Implications, (n.d.). https://openai.com/blog/better-language-models/ (accessed June 3, 2019).

[17] Google Colaboratory, (n.d.). https://colab.research.google.com (accessed June 2, 2019).

[18] N. Shepperd, memory_saving_gradients.py, (n.d.). https://github.com/nshepperd/gpt-2/blob/finetuning/src/memory_saving_gradients.py.

[19] H. Face, The Big-&-Extending-Repository-of-Transformers: Pretrained PyTorch models for Google's BERT, OpenAI GPT & GPT-2, Google/CMU Transformer-XL., (n.d.). https://github.com/huggingface/pytorch-pretrained-BERT (accessed June 3, 2019).

[20] First 100 steps of fine-tuning GPT-2, (n.d.). https://data.mendeley.com/datasets/cgy6ng9kwm/draft?a=c9b6c696-f768-46de-b531-7c2b5479bc50.

[21] J. Vig, Visualizing Attention in Transformer-Based Language Representation Models, (2019). http://arxiv.org/abs/1904.02679 (accessed April 26, 2019).

[22] A. Holtzman, J. Buys, M. Forbes, Y. Choi, The Curious Case of Neural Text Degeneration, (2019). http://arxiv.org/abs/1904.09751 (accessed May 21, 2019).

[23] Unconditional sampling results, (n.d.). https://data.mendeley.com/datasets/wftfn4rs4p/draft?a=009a0411-eb5c-4dcf-bc0d-4995842a38ae.

[24] Conditional sampling results (1), (n.d.). https://data.mendeley.com/datasets/sp3g6c4mc5/draft?a=df172e77-ed9f-4b9d-9339-e1e8305a6d3d.

[25] Conditional sampling results (2), (n.d.). https://data.mendeley.com/datasets/dnxdrgr3h6/draft?a=173df973-966c-4c1f-b3f6-1417d6aa1f4a.

[26]  Z.Li, X.Ding, T.Liu, Story Ending Prediction by Transferable BERT, (2019). http://arxiv.org/abs/1905.07504 (accessed June2, 2019).

[27]  T.Pires, E.Schlinger, D.Garrette, How multilingual is Multilingual BERT?, ArXiv1906.01502v1 [Cs]. (2019). http://arxiv.org/abs/1906.01502v1 (accessed June10, 2019).

## Appendix A

- The following SQL selects the first claims of all US utility patents in 2013 and aggregates the CPC codes at subclass level: (data source: Google Patents Public Datasets on BigQuery)

SELECT STRING_AGG(distinct t2. group_id order by t2. group_id) AS cpc_ids, t1.id, t1.date, text

FROM `patents-public-data.patentsview.patent` t1,

`patents-public-data.patentsview.cpc_current` t2,

`patents-public-data.patentsview.claim` t3

where t1.id = t2.patent_id

and t1.id = t3.patent_id

and timestamp(t1.date) >= timestamp('2013-01-01')

and timestamp(t1.date) <= timestamp('2013-12-31')

and t3.sequence='1'

and t1.type='utility'

group by t1.id, t1.date, t3.text